\documentclass[11pt]{article}

\usepackage[preprint]{acl}
\usepackage{times}
\usepackage{latexsym}
\usepackage[T1]{fontenc}
\usepackage[utf8]{inputenc}
\usepackage{microtype}
\usepackage{inconsolata}
\usepackage{amsmath}
\usepackage{graphicx}
\usepackage{amssymb}
\usepackage{booktabs}
\usepackage{multirow}
\usepackage{array}
\usepackage{tikz}
\usepackage{bbm}
\usetikzlibrary{positioning, shapes, arrows.meta}

\usepackage[table,xcdraw]{xcolor}

\usepackage[most]{tcolorbox}

\usepackage{verbatim}
\usepackage{stfloats}
\usepackage{listings}

\newcounter{promptctr}
\renewcommand{\thepromptctr}{Prompt~\Alph{promptctr}}

\definecolor{examplebg}{HTML}{D1FFBD}
\definecolor{examplebg2}{HTML}{F1F1EE}

\usepackage{enumitem}
\usepackage{mdframed}

\newtcolorbox{exampleblock}{
  enhanced,
  colback=examplebg2,
  colframe=examplebg2,   % same color → invisible border
  boxrule=0pt,
  arc=2mm,
  outer arc=2mm,
  left=8pt,
  right=8pt,
  top=6pt,
  bottom=6pt,
  boxsep=0pt,
  before skip=8pt,
  after skip=8pt,
  sharp corners=all,
  breakable
}

\tcbset{
  bw:domain/.style={
    colback=white,
    colframe=black!40,
    coltitle=black,
    fonttitle=\bfseries\sffamily,   % ✅ fixed
    colbacktitle=examplebg,
    boxrule=0.5pt,
    titlerule=0pt,
    arc=2pt,
    left=4pt,right=4pt,top=4pt,bottom=4pt,
    toptitle=2pt,bottomtitle=2pt,
  }
}

\newcommand{\mytcbinputwide}[5]{% file, title, num, style, label
  \begin{figure*}[t]
  \centering
  \refstepcounter{promptctr}
  \phantomsection
  \begin{tcolorbox}[title={\thepromptctr: #2},#4,width=\textwidth,enhanced]
    \lstinputlisting{#1}
    \label{#5}
  \end{tcolorbox}
  \vspace{-4pt}
  \end{figure*}
}

\newcommand{\promptref}[1]{%
  \hyperref[#1]{\ref*{#1}}%
}

\newcommand{\promptrefp}[1]{%
  \hyperref[#1]{\ref*{#1} (p.~\pageref*{#1})}%
}

\usepackage{fontawesome5}

\title{\textsc{Integrity \faShield* Shield} \\ A System for Ethical AI Use \& Authorship Transparency in Assessments}

\author{
Ashish Raj Shekhar\thanks{contributed equally} \quad
Shiven Agarwal\footnotemark[1] \quad
Priyanuj Bordoloi \quad
Yash Shah \\
\textbf{Tejas Anvekar} \quad
\textbf{Vivek Gupta} \\
Arizona State University \\
\faGlobe~\href{https://shivena99.github.io/IntegrityShield/}{Project Page} \quad
\faPlayCircle~\href{https://shivena99.github.io/IntegrityShield/try}{Demo} \quad
\faVideo~\href{https://www.youtube.com/watch?v=77W_fWW2Agg}{Video} \quad
\faGithub~\href{https://github.com/ShivenA99/IntegrityShield}{Code} \\
\texttt{\{ashekha9, sagar147, pbordolo, yshah124, tanvekar, vgupt140\}@asu.edu}
}

\begin{document}
\maketitle

\begin{abstract}
Large Language Models (LLMs) can now solve entire exams directly from uploaded PDF assessments, raising urgent concerns about academic integrity and the reliability of grades and credentials. Existing watermarking techniques either operate at the token level or assume control over the model's decoding process, making them ineffective when students query proprietary black-box systems with instructor-provided documents.  We present \textsc{Integrity \faShield* Shield}, a document-layer watermarking system that embeds schema-aware, item-level watermarks into assessment PDFs while keeping their human-visible appearance unchanged. These watermarks consistently prevent MLLMs from answering shielded exam PDFs and encode stable, item-level signatures that can be reliably recovered from model or student responses. Across 30 exams spanning STEM, humanities, and medical reasoning, \textsc{Integrity \faShield* Shield} achieves exceptionally high prevention (91-94\% exam-level blocking) and strong detection reliability (89-93\% signature retrieval) across four commercial MLLMs. Our demo showcases an interactive interface where instructors upload an exam, preview watermark behavior, and inspect pre/post AI performance \& authorship evidence. 
\end{abstract}

\section{Introduction}
\label{sec:intro}

\begin{figure}[t]
    \centering
    \includegraphics[width=0.95\linewidth]{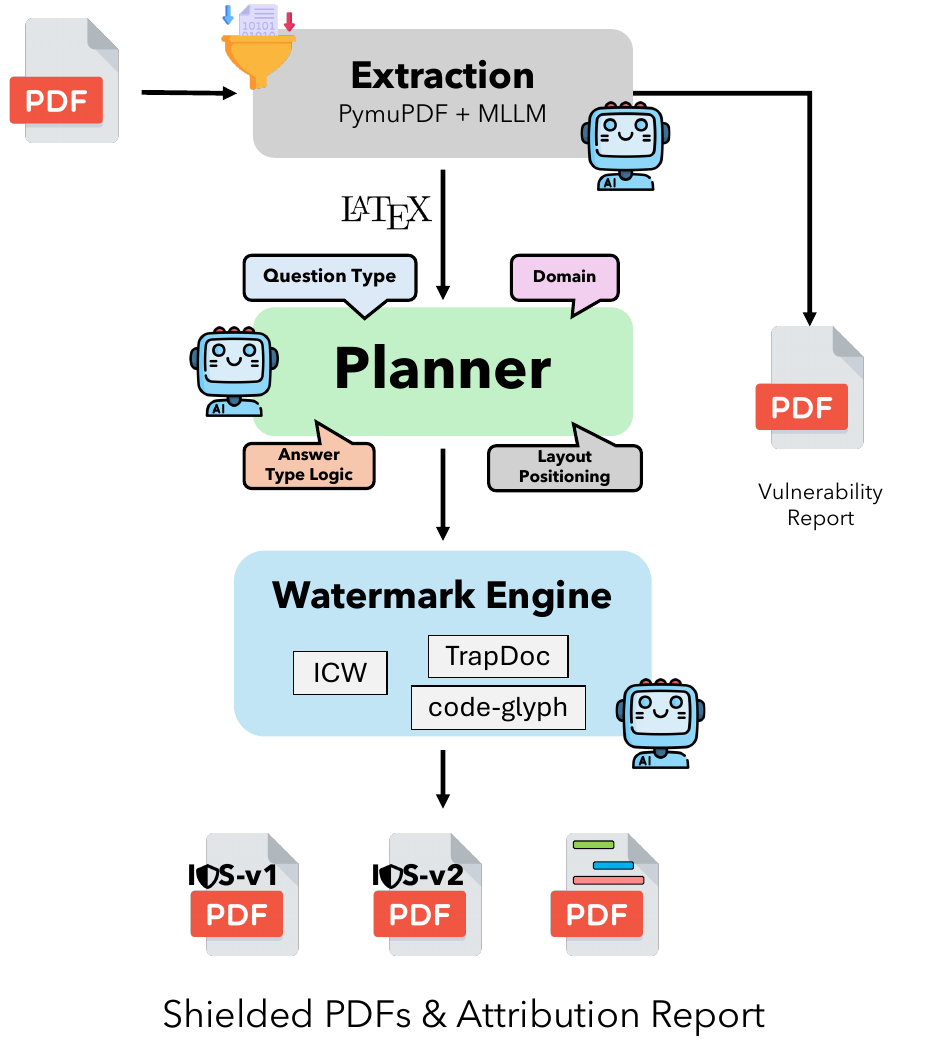}
    \vspace{-0.5em}
    \caption{Overview of \textsc{Integrity \faShield* Shield}. The system extracts question structure from an assessment PDF, uses LLM-based planner to select schema-aware watermarking tactics, \& applies document-layer perturbations through the watermark engine. It outputs shielded PDF variants (\textbf{I\faShield*S-v1}, \textbf{I\faShield*S-v2}) \& an attribution report summarizing AI vulnerability along with authorship signals.}\label{fig:is-arch}
    \vspace{-1.0em}
\end{figure}

LLMs \& MLLMs can now interpret full PDF assessments, reason over diagrams and tables, and produce fluent step-by-step solutions within seconds. While these capabilities expand access to high-quality assistance, they simultaneously undermine the credibility of homework and online exams by enabling students to outsource entire assessments to AI tools \citep{openai_chatgpt,gemini15,susnjak2022chatgptendonlineexam}. 

Institutions have responded with post-hoc detection (e.g., authorship classifiers \citep{pangramtext,thai2025editlensquantifyingextentai}) \& surveillance-heavy proctoring (e.g., keystroke, browser, or gaze monitoring \citep{atoum2017automated,kundu2024keystroke}). However, detectors struggle with short answers, code, paraphrasing, \& mixed authorship \citep{mitchell2023detectgpt,niu2024cheating}, while invasive monitoring raises significant privacy, accessibility, \& equity concerns.

These approaches share a fundamental limitation: they analyze the \emph{student’s output}. In practice, the dominant workflow is the opposite-students upload \emph{instructor-provided PDFs} to black-box AI systems. Existing watermarking methods, which modify generation at the model’s decoder \citep{kirchenbauer2023watermark,liu2025icw}, cannot operate in this setting.

This motivates a new question: \emph{Can assessments themselves be instrumented so that AI reliance becomes observable, without altering visible exam content or student workflows?}

\paragraph{From detection to document-level watermarking.}
We exploit the render-parse gap in PDFs: what humans see often differs from what AI parsers ingest. By injecting invisible text, glyph remappings, and lightweight overlays, we influence model interpretation while leaving the exam visually unchanged. \textsc{Integrity \faShield* Shield} operationalizes this idea as an authorship-aware watermarking system. Rather than asking whether a student cheated, we ask: \emph{to what extent do model-generated responses follow a consistent watermark signature embedded in the exam?} This reframing provides instructors with an interpretable notion of authorship degree while maintaining fairness for honest students.

Finally, we summarize our contributions as:
\begin{itemize}
    \item We introduce \textsc{Integrity \faShield* Shield}, a document-layer watermarking system that embeds schema-aware watermarks into assessment PDFs while keeping their human-visible appearance unchanged.
    \item We develop an LLM-driven planner and PDF watermark engine that adapt tactics to question type,achieving consistently high prevention (91-94\% exam-level blocking) and strong detection reliability (89-93\% retrieval) across four commercial MLLMs on a ten-exam benchmark.
    \item We release an interactive demo that allows instructors to upload exams, preview watermarks, and inspect pre/post AI performance and authorship evidence, enabling ethical and transparent AI use in education.
\end{itemize}

\section{Background and Related Work}
\label{sec:related}

\subsection{AI assistance and mixed authorship}
LLMs increasingly participate in writing and problem-solving tasks, often producing blended human-AI content. Recent work formalizes this as \emph{homogeneous} vs.\ \emph{heterogeneous} mixed authorship \citep{thai2025editlensquantifyingextentai}. Existing detectors-including perplexity-based methods \citep{mitchell2023detectgpt}, style-based classifiers \citep{pangramtext}, and multilingual cheating detectors \citep{niu2024cheating}-struggle with short answers, paraphrasing, and multi-author mixtures, and they analyze only the \textit{output}, leaving the assessment itself uninstrumented.

\subsection{AI watermarking}
Watermarking embeds provenance signals into generated content, typically by modifying decoding distributions \citep{kirchenbauer2023watermark} or via prompt-based in-context cues \citep{liu2025icw}. Parallel work explores invisible watermarks for AI-generated writing that survive paraphrasing and editing \citep{pmlr-v235-liu24e}. These methods assume control over generation, which is infeasible when students query proprietary black-box systems using instructor-provided PDFs.

\subsection{Document-layer perturbations}
Recent work shows that perturbing the PDF substrate-via phantom tokens, font CMaps, or off-page text-can induce systematic model errors without affecting human readability \citep{jin2025trapdoc,xiong2025invisiblepromptsvisiblethreats}. Our work builds on these insights but shifts the objective: rather than deceiving models or detecting cheating, we embed \emph{recoverable watermark signatures} that quantify the extent of AI involvement in solving an exam.

\section{\textsc{Integrity \faShield* Shield}  \\ System Architecture \& Workflow}
\label{sec:system}
\subsection{Design Principles}

\textsc{Integrity \faShield* Shield}  is designed as a practical tool for instructors who want to harden PDF-based assessments against AI assistants without redesigning their exams or changing grading workflows. The system keeps all human-facing content (layout, typography, pagination, item numbering) unchanged while embedding signals that reliably influence model-side parsing. It adapts watermark tactics to the item schema, treating \textbf{MCQ}, \textbf{true/false}, and \textbf{Long-Form} questions differently; remains robust across black box MLLMs; and exposes a lightweight web interface where instructors can upload assessments, preview watermark behavior, and inspect calibration and authorship signals with minimal configuration.

\subsection{End-to-End Architecture and Workflow}

\autoref{fig:is-arch} summarizes the end-to-end workflow of \textsc{Integrity \faShield* Shield} . A single-page web front end communicates with a stateless backend that operates directly on the PDF substrate. The backend is organized around four logical services: document ingestion, which parses the uploaded PDF into a structured item schema with stems, options, diagrams, and answer keys; strategy planning, via a lightweight instruction-tuned LLM that assigns a watermark plan to each item based on its type, gold answer, and local layout metadata; PDF rewriting, which applies the plan while enforcing that the rendered appearance of the document remains unchanged; and an authorship and calibration service that runs reference models on original and watermarked PDFs and later scores submitted answers against stored watermark signatures.

From an instructor’s perspective, this architecture is depicted in three-stage interaction flow.

\begin{figure}[!ht]
    \centering
    \includegraphics[width=0.95\linewidth]{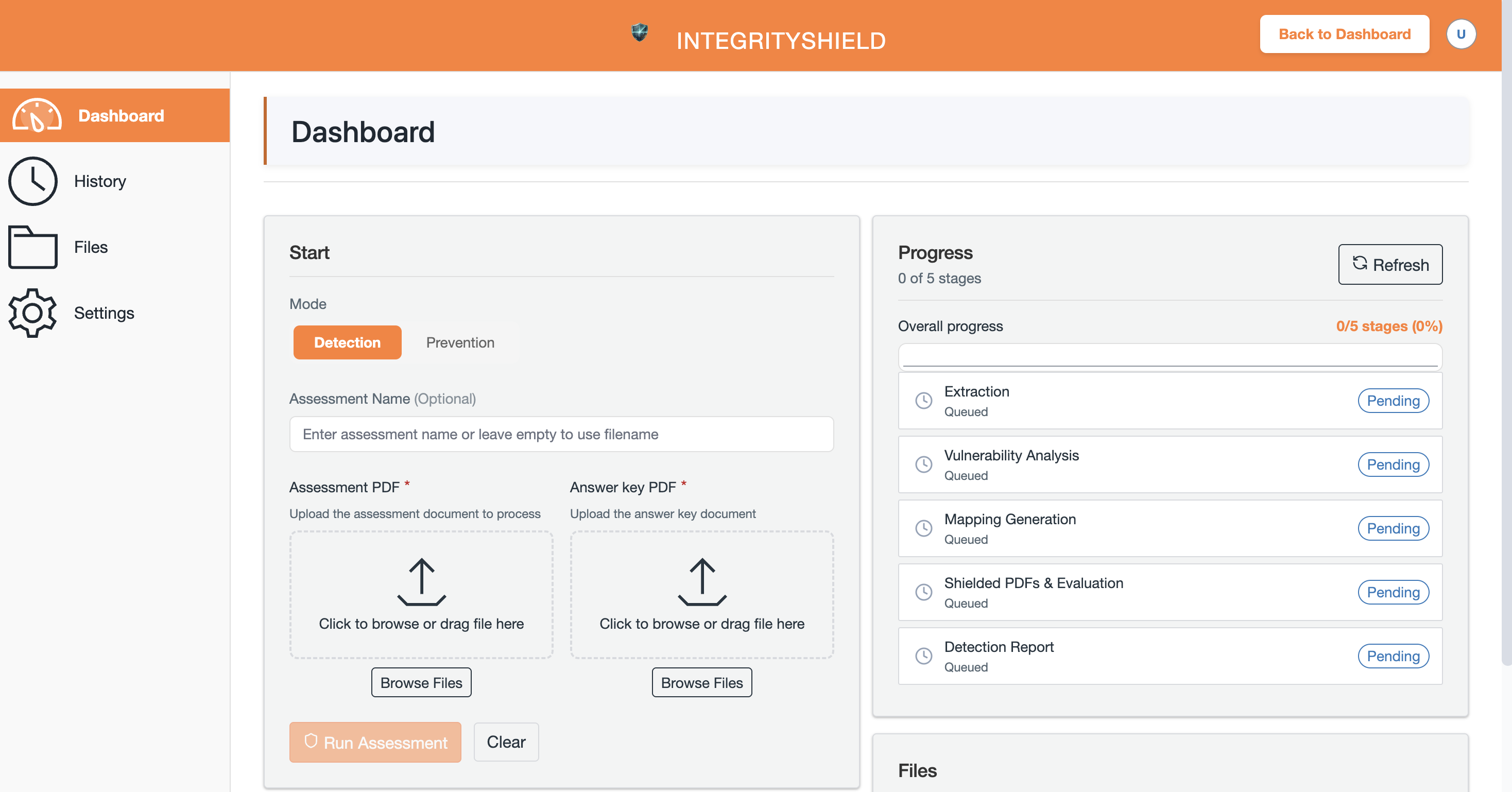}
    \caption{Stage~1: Upload \& Watermark Planning. Instructors upload an assessment PDF \& answer key, after which the system extracts question structure \& previews the planned schema-aware watermarking strategies.
    \vspace{-1.0em}
}\label{fig:stage1}
\end{figure}

\paragraph{Stage 1: Upload \& Watermark Planning.} The instructor uploads an exam PDF through the browser. The ingestion service segments pages into questions, detects answer options \& numbering, \& associates inline figures or tables with the relevant items, producing a compact item schema with content spans, page coordinates, and answer keys when available, best depicted in \autoref{fig:stage1}. The strategy planner then assigns, for each item, either a \emph{target distractor} (for multiple-choice and true/false questions) or a small set of signature keyphrases (for long-form questions), and decides which document-layer mechanisms to apply. The interface presents a split-screen preview of original and watermarked pages with per-question summaries of the chosen strategy, allowing instructors to inspect and optionally disable aggressive tactics (such as strong glyph remapping) before proceeding.

\begin{figure}[!ht]
    \centering
    \includegraphics[width=0.95\linewidth]{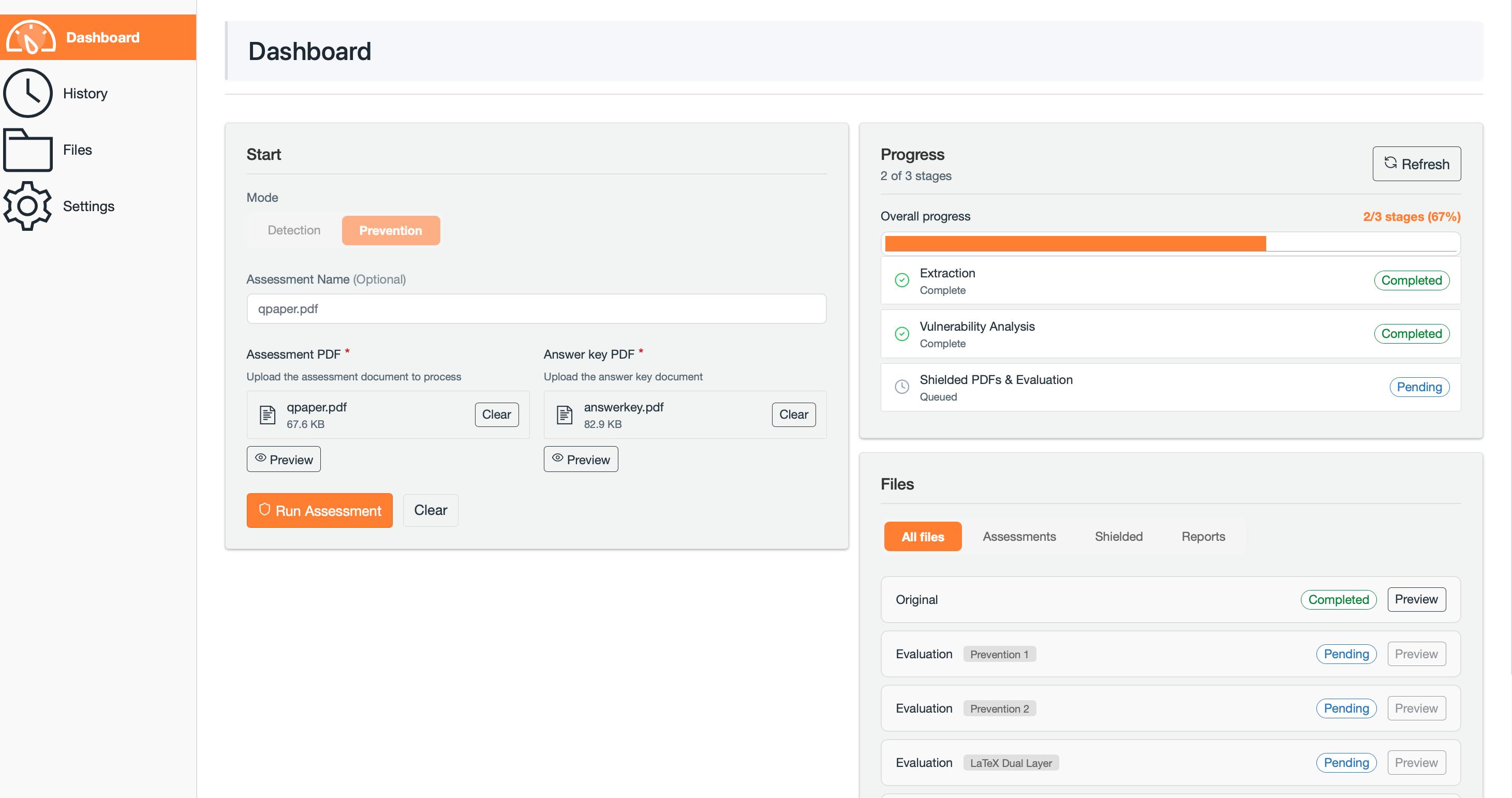}
    \caption{Stage~2: Watermark Embedding \& AI Calibration. 
After planning, the system applies document-layer watermarks to the assessment PDF and evaluates original vs.\ watermarked versions against multiple MLLMs to generate prevention and detection reports.}\label{fig:stage2}
\vspace{-1.0em}
\end{figure}
\paragraph{Stage 2: Watermark Embedding \& AI Calibration.}
Once the plan is confirmed, the PDF rewriting service applies it directly to the assessment file. It injects invisible text spans anchored near stems and options, applies CMap-based glyph remapping so that visually identical tokens are parsed differently by models, and, when appropriate, adds clipped or off-page overlays that insert distractor-oriented cues into the parseable stream while keeping them outside the visible canvas, best depicted in \autoref{fig:stage2}. 
We instantiate two watermark configurations, \textbf{I\faShield*S-v1} and \textbf{I\faShield*S-v2}, which differ in the density and combination of these mechanisms: \textbf{I\faShield*S-v1} uses a lighter mix of hidden-text and minimal glyph remapping, whereas \textbf{I\faShield*S-v2} employs stronger multi-layer perturbations for maximal robustness across parsing pipelines. 
After rewriting, the system verifies that the rendered appearance of the PDF matches the original across common viewers (Adobe Reader, Chrome, macOS Preview). In the same stage, the authorship and calibration service evaluates both the original and watermarked versions with a panel of reference models in a simulated \emph{``student uploads the exam''} setting, computing pre- versus post-watermark accuracy, the fraction of incorrect answers that land on intended distractors, and per-item watermark retrieval rates. An interactive report summarizes these statistics and assists instructors in selecting an appropriate watermark \emph{``strength''} preset.

\begin{figure}[!ht]
    \centering
    \includegraphics[width=0.95\linewidth]{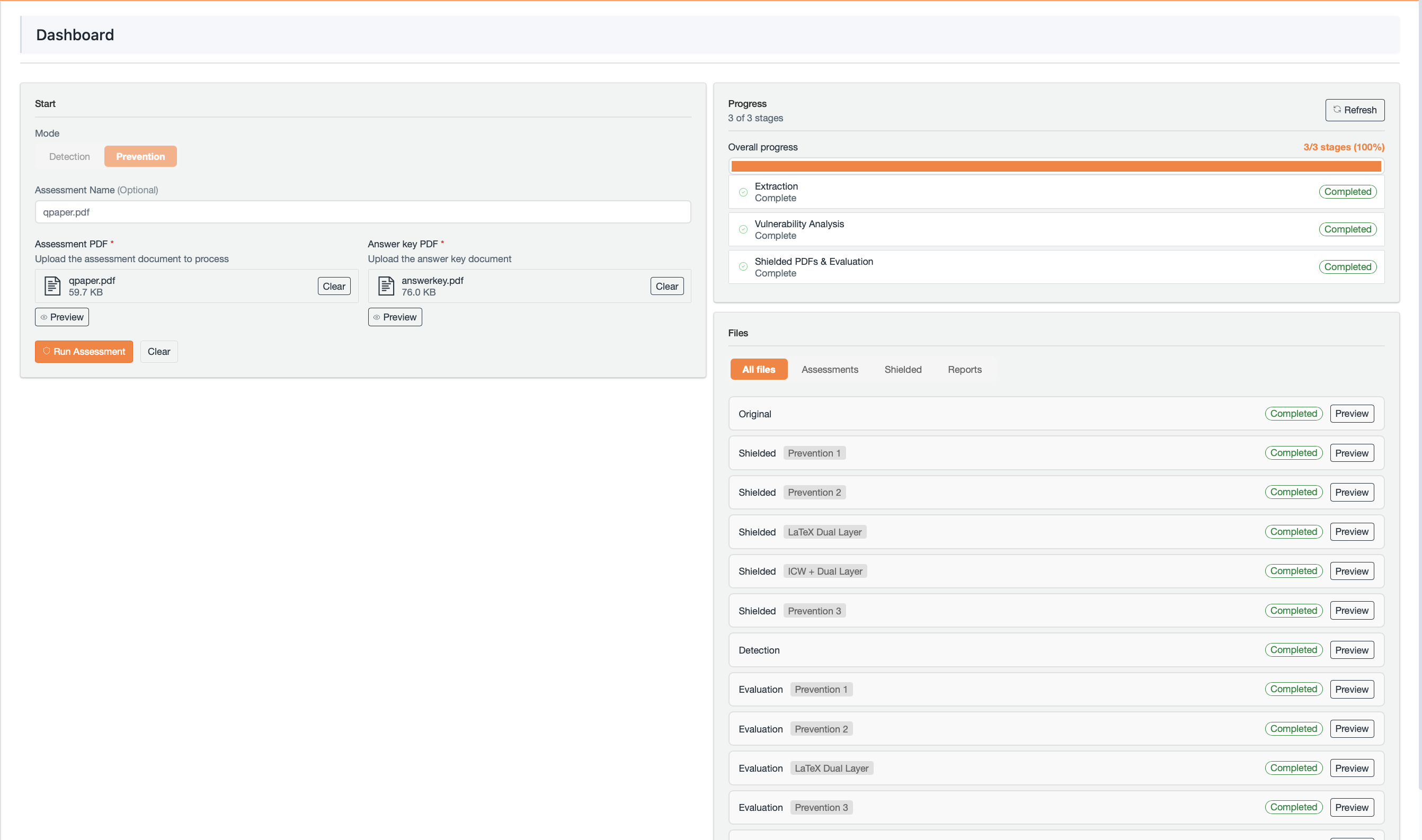}
    \caption{Stage~3: Authorship Analysis. The dashboard displays per-question watermark retrieval, exam-level authorship scores, and previewable shielded PDFs, enabling instructors to inspect AI-reliance signals.}\label{fig:stage3}
    \vspace{-1.0em}
\end{figure}

\paragraph{Stage 3: Authorship Analysis.}
After an assessment has been protected with our \textbf{I\faShield*S} watermarked PDFs, instructors can use it to analyze responses. The interface accepts either raw model outputs (for research) or, can also anonymized student responses exported from a learning management system, best depicted in \autoref{fig:stage3}. For each question, the authorship engine checks whether the response follows the stored watermark signature: for objective questions, this reduces to matching the target distractor (or a small tied set); for long-form questions, a judge LLM scores how closely the response tracks the watermark’s keyphrases or erroneous reasoning patterns. These per-item scores are aggregated into an exam-level authorship degree and displayed on a dashboard with cohort-level distributions and drill-down views for individual questions. The tool is explicitly positioned as an aid for triage rather than an automatic decision system: high authorship scores are intended to trigger follow-up actions such as brief oral checks or additional written assessments, keeping human judgment in the loop.

\section{Experiments}
\label{sec:experiments}

\subsection{Experimental Setup}

\paragraph{Models and prompts.}
We evaluate \textsc{Integrity \faShield* Shield} against a panel of four proprietary frontier MLLMs that support direct PDF ingestion: GPT-5, Claude Sonnet-4.5, Grok-4.1, and Gemini-2.5 Flash.  All models are treated as black boxes and accessed via their official APIs with temperature set to $0$ and maximum output length sufficient to cover all questions in an exam. For each exam, we use a minimal, instruction-style prompt that (i) asks the model to answer all questions in order, (ii) returns a structured list of answers (e.g., \emph{``Q1: A, Q2: C, \dots''} for MCQ and T/F; numbered paragraphs for Long-Form (LF)), and (iii) forbids external tools or web browsing.
We use the same prompting templates for original and watermarked PDFs; full prompt text for MCQ, T/F, and LF questions appears in Appendix~\ref{sec:appendix_prompts} as \promptref{prompt:MCQ}, \promptref{prompt:TF}, and \promptref{prompt:LF}.

\paragraph{Baselines.}
We compare our two watermark configurations, \textbf{I\faShield*S-v1} and \textbf{I\faShield*S-v2}, against three document- or prompt-level baselines.
\textbf{ICW} is an in-context watermarking method that attempts to steer model outputs using prompt-side patterns without modifying the PDF content using invisible white color small sized font size ~(0.1-0.5) .~\citep{liu2025icw}.
\textbf{\texttt{code-glyph}} is a document-layer baseline that manipulates bitcode-to-glpyh mapping on question text to perturb parsing while keeping human readability intact\citep{xiong2025invisiblepromptsvisiblethreats}.
\textbf{\textsc{TrapDoc}} adapts document-layer perturbations that introduce phantom tokens and layout tricks to cause models to produce plausible but incorrect answers without visible changes to the PDF~\citep{jin2025trapdoc}.
In contrast, \textbf{I\faShield*S-v1} and \textbf{I\faShield*S-v2} operate directly at the PDF substrate with schema-aware hidden text, glyph remapping, and overlays; \textbf{I\faShield*S-v1} applies a lighter combination aimed at minimal perturbation, while \textbf{I\faShield*S-v2} uses denser, multi-layer perturbations for maximal robustness.

\paragraph{Benchmark dataset.}
To approximate real assessment settings, we compile a diverse benchmark of exam-style PDFs by web-scraping publicly available quizzes, homework sets, and midterm assessments from university course websites (e.g., Stanford and other institutions). The collected material spans mathematics, science, programming, humanities, and medical reasoning, and includes a mix of MCQ, T/F, and long-form questions. From this pool, we sample $\approx$10\% of  items to construct our benchmark, selecting documents that (i) contain at least ten questions, (ii) include at least three question formats, and (iii) render cleanly as PDFs. All items and answer keys are qualitatively reviewed by two authors and spot-validated quantitatively (e.g., via official solutions when available) to filter out ambiguous or mislabeled questions.

\paragraph{Evaluation metrics.}
We evaluate \textsc{Integrity \faShield* Shield} along two complementary axes: \emph{prevention}, which measures how strongly watermarking disrupts a model's ability to answer correctly, \& \emph{detection}, which captures how reliably watermark signatures can be recovered from model or student responses.

For \textbf{prevention}, we simply check whether watermarking causes the model to fail or refuse to answer the exam PDF. For exam $d$, $\text{Prev}(f,d)=1[y^{\text{wm}} \ \text{contains no usable answers}],$ and we report the percentage of PDFs where this occurs.

For \textbf{detection}, we measure the degree to which model outputs follow the embedded watermark signature. For each item, the authorship engine assigns a retrieval score $c_i \in [0,1]$: for MCQ; T/F, $c_i = 1$ iff the model selects the target distractor; for long-form, $c_i$ is produced by a judge LLM evaluating alignment with watermark keyphrases. The exam-level detection score is
\[
\text{Det}(d',y) = \frac{1}{n_{d'}} \sum_{i=1}^{n_{d'}} c_i,
\]
representing the proportion of responses that exhibit watermark-consistent behavior. We report detection scores per model and method, with breakdowns by question type.

\subsection{Quantitative Analysis on Prevention and Detection}
\begin{table}[!ht]
\small
\centering
\begin{tabular}{lcccc}
\hline \hline
\textbf{Method} & \textbf{GPT} & \textbf{Claude} & \textbf{Grok} & \textbf{Gemini} \\
\hline
\multicolumn{5}{c}{\cellcolor[HTML]{D5FFD4}\textit{\textbf{Prevention-ASR}}} \\
\textbf{ICW}             & 07.20 & 05.80 & 04.10 & 03.50 \\
\textbf{\texttt{code-glyph}}      & 86.30 & 84.7 & 83.20 & 81.90 \\
\textbf{\textsc{TrapDoc}}& 88.70 & 86.40 & 85.10 & 40.50 \\
\textbf{I\faShield*S-v1} & 91.20 & 90.80 & 90.50 & 90.10 \\
\textbf{I\faShield*S-v2} & \textbf{93.60} & \textbf{92.90} & \textbf{92.30} & \textbf{91.70} \\
[0.8ex]
\multicolumn{5}{c}{\cellcolor[HTML]{FAE4CA}\textit{\textbf{Detection}}} \\
\textbf{ICW}            & 06.80 & 05.30 & 04.60 & 03.20 \\
\textbf{\texttt{code-glyph}}      & 85.90 & 84.20 & 82.70 & 81.40 \\
\textbf{\textsc{TrapDoc}}& 87.90 & 85.80 & 84.60 & 43.40 \\
\textbf{I\faShield*S-v1} & 90.70 & 90.30 & 89.90 & 89.50 \\
\textbf{I\faShield*S-v2} & \textbf{92.80} & \textbf{92.10} & \textbf{91.60} & \textbf{91.00} \\
\hline \hline
\end{tabular}
\caption{
Prevention and detection performance across models.  Prevention-ASR is the percentage of exam PDFs on which watermarking causes the model to refuse or fail to produce usable answers. Detection is the percentage of questions whose responses follow the embedded watermark signature. For both metrics, higher is better. ICW: in-context watermarking; \texttt{code-glyph}: glyph perturbation; \textsc{TrapDoc}: phantom-token PDF attack; IS: \textsc{Integrity \faShield* Shield} variants.
}\label{tab:prevention_detection} \vspace{-0.5em}
\end{table}

\autoref{tab:prevention_detection} summarizes prevention and detection performance for all baselines and our \textbf{I\faShield*S} variants across GPT, Claude, Grok, and Gemini.

In the \emph{Prevention-ASR} block, ICW almost never prevents models from answering full exams, with single-digit prevention rates across all models. This confirms that prompt-only steering is ineffective when students upload raw PDFs to black-box MLLMs. Document-layer baselines such as \texttt{code-glyph} and \textsc{TrapDoc} are substantially stronger on GPT, Claude, and Grok (around 83--89\% prevention), but \textsc{TrapDoc} degrades sharply on Gemini (40.5\%), suggesting that its perturbations do not transfer reliably across parsing and model stacks. By contrast, \textbf{I\faShield*S-v1} and \textbf{I\faShield*S-v2} achieve consistently high prevention on \emph{all} models (90--94\%), indicating that schema-aware, multi-layer PDF watermarking can robustly block end-to-end exam solving for contemporary MLLMs.

The \emph{Detection} block shows a similar pattern. ICW again yields negligible detection rates, while \texttt{code-glyph} and \textsc{TrapDoc} achieve strong detection on GPT, Claude, and Grok (mid--80s), but \textsc{TrapDoc} drops to 43.4\% on Gemini. In contrast, \textbf{I\faShield*S-v1} and especially \textbf{I\faShield*S-v2} maintain high detection performance across all four models (around 89--93\%), meaning that whenever models do attempt to answer on watermarked exams, their outputs follow the embedded watermark signatures in a highly consistent way, enabling reliable authorship attribution.

\subsection{I\faShield*S Performance for Question-Category}

\begin{table}[!ht]
\centering
\resizebox{\linewidth}{!}{%
\begin{tabular}{lcccccccc}
\hline\hline
\multirow{2}{*}{\textbf{Type}} &
\multicolumn{2}{c}{\textbf{GPT}} &
\multicolumn{2}{c}{\textbf{Claude}} &
\multicolumn{2}{c}{\textbf{Grok}} &
\multicolumn{2}{c}{\textbf{Gemini}} \\
& $w/o$ & $w/$ & $w/o$ & $w/$ & $w/o$ & $w/$ & $w/o$ & $w/$ \\
\hline
MCQ        & 96.2 & \cellcolor[HTML]{D4E6F1}\textbf{7.8} & 95.8 & \cellcolor[HTML]{D4E6F1}\textbf{6.9} & 94.9 & \cellcolor[HTML]{D4E6F1}\textbf{5.7} & 94.1 & \cellcolor[HTML]{D4E6F1}\textbf{4.3} \\
T/F        & 95.7 & \cellcolor[HTML]{D4E6F1}\textbf{6.5} & 95.3 & \cellcolor[HTML]{D4E6F1}\textbf{5.8} & 94.6 & \cellcolor[HTML]{D4E6F1}\textbf{4.9} & 93.8 & \cellcolor[HTML]{D4E6F1}\textbf{3.6} \\
LF         & 96.8 & \cellcolor[HTML]{D4E6F1}\textbf{5.2} & 96.4 & \cellcolor[HTML]{D4E6F1}\textbf{4.6} & 95.3 & \cellcolor[HTML]{D4E6F1}\textbf{3.8} & 94.7 & \cellcolor[HTML]{D4E6F1}\textbf{3.1} \\
\hline\hline
\end{tabular}
}
\caption{
Per-question-type answer accuracy without ($w/o$) and with ($w/$) our \textsc{Integrity Shield}~\faShield* watermarks. 
Values show the residual accuracy of each model on shielded exams; lower $w/$ accuracy indicates stronger prevention for that question type.}\label{tab:detection_split} \vspace{-0.5em}
\end{table}

\autoref{tab:detection_split} breaks down the impact of \textbf{IS} on answer accuracy by question type (MCQ, T/F, LF) and model, comparing performance on original ($w/o$) and watermarked ($w/$) exams. Without watermarking, all four MLLMs attain very high accuracy across categories (typically 94-97\%), reflecting their strong baseline performance on our exam-style benchmark.

With \textsc{Integrity \faShield* Shield} enabled, residual accuracy collapses into the low single digits for every model and question type (3-8\%), corresponding to an 85-90 point drop.
Long-form questions show the largest reductions for GPT and Claude, while MCQ and T/F items are also heavily disrupted across all models.
These results indicate that our document-layer watermarks are effective not only at the exam level, but also uniformly across different assessment formats.

\subsection{Utility of \textsc{Integrity \faShield* Shield}}
\begin{table}[!ht]
\centering
\small
\resizebox{\linewidth}{!}{
\begin{tabular}{lcccc}
\hline\hline
\textbf{Attack Method} &
\textbf{GPT} &
\textbf{Claude} &
\textbf{Grok} &
\textbf{Gemini} \\
\hline

\rowcolor[HTML]{F8D7DA}
\textbf{ICW} &
A  &
A  &
A  &
A  \\

\rowcolor[HTML]{F8D7DA}
\textbf{\texttt{code-glyph}} &
A  &
A  &
A  &
A \\

\rowcolor[HTML]{F8D7DA}
\textbf{\textsc{TrapDoc}} &
A  &
A &
A &
A \\

% ---------------- OUR METHODS ----------------
\rowcolor[HTML]{D5FFD4}
\textbf{I\faShield*S-v1} &
B &
B &
B &
B \\

\rowcolor[HTML]{D5FFD4}
\textbf{I\faShield*S-v2} &
C &
C &
C &
C \\

\hline\hline
\end{tabular}
}\caption{Model predictions across attack methods for the OSI model question. 
\textbf{\textit{Q:}} \textit{Which layer of the OSI model is responsible for routing packets between networks?}
\textbf{\emph{Gold Answer: B}}}
\label{tab:osi_attacks}
\vspace{-1.0em}
\end{table}

Beyond aggregate metrics, \textsc{Integrity \faShield* Shield} provides instructors with actionable, item-level evidence of AI reliance. 
\autoref{tab:osi_attacks} illustrates this with a qualitative example: on an OSI-model question, all baseline attacks (\texttt{ICW}, \texttt{code-glyph}, \textsc{TrapDoc}) collapse to the same incorrect prediction across models, offering no consistent signal for attribution. In contrast, our schema-aware variants (\textbf{I\faShield*S-v1}, \textbf{I\faShield*S-v2}) drive models toward distinct, watermark-aligned distractors (B and C, respectively), enabling clear and separable authorship signatures.

In a \emph{prevention-focused} deployment, the system summarizes where watermarking fully blocks a model from answering an exam, providing a document-level view of which assessments are resilient to AI-based shortcuts. In a \emph{detection-focused} deployment, the system aggregates authorship evidence across questions, showing, for example, that \emph{``Q3 follows the} I\faShield*S-v2 \emph{signature across multiple models''}. 

These reports are intended as triage tools: instructors can identify items likely influenced by AI, perform brief oral checks or follow-up tasks, and intervene proportionally. By surfacing interpretable authorship signals rather than relying on opaque classifiers or intrusive proctoring \textsc{Integrity \faShield* Shield} enables ethical, transparent, and governance-aligned AI use in educational assessments.

\section{Conclusion}
\label{sec:conclusion}

\textsc{Integrity \faShield* Shield} demonstrates that assessment integrity can be strengthened without invasive monitoring by instrumenting the exam document itself. By operating directly at the PDF substrate, our system embeds schema-aware watermarks that both (i) prevent modern MLLMs from answering shielded exams (91-94\% exam-level blocking) \& (ii) yield stable, recoverable authorship signatures (89-93\% retrieval) when AI is used. These effects hold consistently across question types \& four commercial MLLMs, highlighting the robustness of document-layer watermarking as a practical defense.

The demo showcases a complete workflow for real instructional use: uploading an exam, previewing watermark strategies, generating shielded variants, running automated AI calibration, and inspecting item-level authorship evidence. This combination of prevention and attribution provides instructors and institutions with actionable, interpretable signals? supporting fair assessment practices, targeted follow-up, and transparent communication with students.

We hope \textsc{Integrity \faShield* Shield} serves as a step toward ethically grounded AI governance in education, enabling institutions to observe AI reliance without resorting to surveillance or restricting access to assistive technologies.

\section{Limitations}
\label{sec:limitations}

Our evaluation is limited to a ten-exam benchmark, a fixed set of frontier MLLMs, and simulated usage in which models directly consume instructor PDFs. Real-world deployments may involve broader variation in domains, languages, accessibility workflows (e.g., screen readers), and institution-specific formats. As MLLMs and their PDF-parsing pipelines evolve, watermark robustness may drift, necessitating periodic re-calibration.

\textsc{Integrity \faShield* Shield} is not a definitive detector of misconduct. Authorship scores indicate alignment with watermark signatures-not whether a student violated policy-and should be used as a triage signal for human follow-up (e.g., brief oral checks), not as automatic evidence for sanctions.

Finally, our approach assumes institutional control over assessment PDFs. Similar watermarking techniques could be misapplied to non-assessment documents, so we explicitly restrict the intended use to formal educational settings with clear governance, transparency, \& AI-use policies.

\section{Ethics Statement}

This work aims to support ethical and transparent AI use in educational assessment settings. 
\textsc{Integrity \faShield* Shield} operates exclusively on instructor-provided documents and does not monitor students, avoiding surveillance-heavy practices such as keystroke logging, webcam tracking, or device control. The system is designed to keep all responses and analyses within institutional infrastructure, respecting student privacy and data-governance requirements.

Authorship scores produced by our watermarking framework indicate alignment with embedded watermark signatures; they do \emph{not} constitute evidence of misconduct. We recommend that institutions (i) clearly communicate AI-use policies and the presence of watermarking to students, (ii) treat high authorship scores only as signals for human review (e.g., follow-up questions or oral checks), and (iii) ensure that any use of these signals aligns with local policies, academic integrity guidelines, and privacy regulations.

All experiments were conducted with fixed model parameters (e.g., temperature, $top_p$, $top_k$) to mitigate stochastic variability in black-box LLMs. Models used in this work (e.g., GPT-5, Gemni-2.5 Flash, Grok-4.1, Claude Sonnet-4.5) were accessed in accordance with their respective usage policies. Data labeling and verification were performed by author-annotators, and AI-based tools (e.g., Grammarly, ChatGPT) were used strictly for language refinement. To the best of our knowledge, this study introduces no additional ethical risks beyond those common to LLM evaluation in controlled educational settings.

\bibliography{custom}

\appendix

\section{Prompts Details}
\label{sec:appendix_prompts}
\mytcbinputwide{prompts/MCQ_Perturbation.tex}{MCQ Perturbation}{0}{bw:domain}{prompt:MCQ}

\mytcbinputwide{prompts/TF_Perturbation.tex}{True False Perturbation}{0}{bw:domain}{prompt:TF}

\mytcbinputwide{prompts/LongForm_Perturbation.tex}{LongForm Perturbation}{0}{bw:domain}{prompt:LF}

\end{document}